\def\year{2021}\relax
\documentclass[letterpaper]{article} 
\usepackage{aaai21}  
\usepackage{times}  
\usepackage{helvet} 
\usepackage{courier}  
\usepackage[hyphens]{url}  
\usepackage{graphicx} 
\usepackage{amssymb}
\usepackage{natbib}  
\usepackage{caption} 
\usepackage[switch]{lineno}
\usepackage{subfigure}
\usepackage{color}
\title{Large Motion Video Super-Resolution with Dual Subnet and Multi-Stage Communicated Upsampling}
\author {
	Hongying Liu\textsuperscript{\rm 1},
	Peng Zhao\textsuperscript{\rm 1},
	Zhubo Ruan\textsuperscript{\rm 1},
	Fanhua Shang\textsuperscript{\rm 1,2}\thanks{Corresponding Author: fhshang@xidian.edu.cn.},
	Yuanyuan Liu\textsuperscript{\rm 1} \\
}
\affiliations {
	\textsuperscript{\rm 1} Key Laboratory of Intelligent Perception and Image Understanding of Ministry of Education, School of Artificial Intelligence, Xidian University;
	\textsuperscript{\rm 2} Peng Cheng Lab, Shenzhen.
}
\begin{document}

\maketitle

\begin{abstract}
Video super-resolution (VSR) aims at restoring a video in low-resolution (LR) and improving it to higher-resolution (HR). Due to the characteristics of video tasks, it is very important that motion information among frames should be well concerned, summarized and utilized for guidance in a VSR algorithm. Especially, when a video contains large motion, conventional methods easily bring incoherent results or artifacts. In this paper, we propose a novel deep neural network with Dual Subnet and Multi-stage Communicated Upsampling (DSMC) for super-resolution of videos with large motion. We design a new module named U-shaped residual dense network with 3D convolution (U3D-RDN) for fine implicit motion estimation and motion compensation (MEMC) as well as coarse spatial feature extraction. And we present a new Multi-Stage Communicated Upsampling (MSCU) module to make full use of the intermediate results of upsampling for guiding the VSR. Moreover, a novel dual subnet is devised to aid the training of our DSMC, whose dual loss helps to reduce the solution space as well as enhance the generalization ability. Our experimental results confirm that our method achieves superior performance on videos with large motion compared to state-of-the-art methods.
\end{abstract}

\section{Introduction}

Video super-resolution (VSR) aims at recovering the corresponding high-resolution (HR) counterpart from a given low-resolution (LR) video \cite{liu2020video}. As an important computer vision task, it is a classic ill-posed problem. In recent years, due to the emergence of 5G technology and the popularity of high-definition (HD) and ultra high-definition (UHD) devices \cite{liu2020single}, the VSR technology has attracted more attention from researchers and has become one of the research spotlights. Traditional super-resolution (SR) methods mainly include interpolation, statistical methods and sparse representation methods.

In recent years, with the rapid development of deep neural networks, deep-learning-based VSR has attracted more attention among researchers. Due to the powerful data fitting and feature extraction ability, such algorithms are generally superior to traditional super-resolution techniques. The first deep-learning-based single image super-resolution (SISR) algorithm is SRCNN \cite{dong2015image}, while the first deep-learning-based VSR algorithm is Deep-DE \cite{liao2015video}. Since then, many deep-learning-based VSR algorithms have been proposed, such as VSRnet \cite{kappeler2016video}, 3DSRNet \cite{kim20183dsrnet}, RBPN \cite{haris2019recurrent} and TDAN \cite{TDAN}. It may be considered that VSR can be achieved by using SISR algorithms frame by frame. However, SISR algorithms ignore temporal consistency between frames and easily brings artifact and jam, leading to worse visual experience. In contrast, VSR methods are usually able to process such consecutive frames and generate HR video with more natural details and less artifact.

There are many VSR methods based on motion estimation and motion compensation (MEMC). They rely heavily on optical flow estimation for consecutive frames, and thus execute compensation to import temporal information to the center frame. For example, SOF-VSR \cite{wang2018learning} proposed a coarse-to-fine CNN to gradually estimate HR optical flow in three stages. And RBPN \cite{haris2019recurrent} presented a back projection module which is composed of a sequence of consecutive alternate encoders and decoders after implicit alignment. They work well for videos consisting of scenes with small motion in a short time, as the optical flow estimation is accurate under such ideal scenes. However, in actual multimedia scenes, the motion are always diverse in different amplitudes. Especially when real-time shooting scenes like extreme sports have been popular, wearable shooting equipments can be widely used and often bring about video jitter. The jitter can easily bring about large motions. In visual tasks, large motion is always based on the consistency assumption of optical flow calculation \cite{gibson1957optical}. If the target motion changes too fast relative to the frame rate, this motion can be called large motion in videos.

Moreover, some VSR methods do not perform explicit MEMC. They directly input multiple frames for spatio-temporal feature extraction, fusion and super-resolution, thus achieve implicit MEMC. For example, 3DSRNet \cite{kim20183dsrnet} and FSTRN \cite{li2019fast} utilize 3D convolution (C3D) \cite{ji20123d} to extract spatio-temporal correlation on the spatio-temporal domain. However, high computational complexity of C3D limits them to develop deeper structures. This probably results in their limited modeling and generalization ability, and the difficulty to adapt to videos with large motion.

To address the above challenges, we propose a novel video super-resolution network with Dual Subnet and Multi-stage Communicated Upsampling (DSMC) to maximize the communication of various decisive information for videos with large motion. DSMC receives a center LR frame and its neighboring frames for each SR. After coarse-to-fine spatial feature extraction on the input frames, a U-shaped residual dense network with 3D convolution (U3D-RDN) is designed for DSMC. It can encode the input features and achieve both fine implicit MEMC and coarse spatial feature extraction on the encoding space as well as reducing the computational complexity. Then U3D-RDN decodes the features by a sub-pixel convolution upsampling layer. After another fine spatial feature extraction, a Multi-Stage Communicated Upsampling (MSCU) module is proposed to decompose an upsampling into multiple sub-tasks. It conducts feature correction with the help of the VSR result from each sub task, and thus makes full use of the intermediate results of upsampling for VSR guidance. Finally, a dual subnet is presented and used to simulate degradation of natural image, and the dual loss between the degraded VSR result and the original LR frame is computed to aid the training of DSMC.

The main contributions of this paper are as follows:

\begin{itemize}
\item We propose a DSMC network for super-resolution of videos with large motion, which is designed to maximize the communication of various decisive information in VSR process and implicitly capture the motion information. Our DSMC can guide the upsampling process with more sufficient prior knowledge than other state-of-the-art ones by the proposed MSCU model. Meanwhile, the proposed U3D-RDN module can learn coarse-to-fine spatio-temporal features from the input video frames, and therefore effectively guide VSR process for large motion.
\item We propose a dual subnet for our DSMC, which can simulate natural image degradation to reduce the solution space, enhance the generalization ability and help DSMC for better training.
\item Extensive experiments have been carried out to evaluate the proposed DSMC. We compare it with several state-of-the-art methods including optical-flow-based and C3D-based ones. Experimental results confirm that DSMC is effective for videos with large motion as well as for generic videos without large motion.
\item Ablation study for each individual design has been conducted to investigate the effectiveness of our DSMC. We can find that MSCU has the greatest influence on the performance as it can recover more details through multi-stage communication. U3D-RDN is also effective for extracting motion information. The ablation study also indicates that the loss functions in dual subnet influence the training of DSMC when the original loss function is under different combinations of Cb and perceptual losses.
\end{itemize}

\section{Related Work}

\subsection{SISR Methods Based on Deep Learning}

Recently, with the development of deep learning, super-resolution algorithms based on deep learning usually perform much better than traditional methods in terms of various evaluation indicators, such as PSNR and SSIM. The first deep-learning-based SISR algorithm (called SRCNN) was proposed by \citet{dong2015image}. It consists of three convolutional layers and learns a non-linear mapping from LR images to HR images by an end-to-end manner. Since then, many deep learning methods have been transferred to SISR, which help subsequent methods obtain greater performance.

Inspired by VGG \cite{simonyan2014very}, some methods generally adopt deeper network architecture, such as VDSR \cite{kim2016accurate}, EDSR \cite{lim2017enhanced} and RCAN \cite{zhang2018image}. However, these methods may suffer from gradient vanishment problem. Therefore, many algorithms such as RDN \cite{zhang2018residual} introduce the skip connection between different layers inspired by the residual network (ResNet) \cite{he2016deep}. In addition, the input size of SRCNN is the same as ground truth, which can lead to a high computational complexity. Therefore, most subsequent algorithms adopt a single LR image as input and execute upsampling on it at the end of the network, such as ESPCN \cite{shi2016real} and DRN \cite{guo2020closed}. Besides, other strategies such as attention mechanism \cite{mnih2014recurrent}, non-local \cite{wang2018non} and dense connection \cite{huang2017densely} are also introduced to enhance the performance of SISR methods.

\begin{figure*}[ht]
  \centering
  \includegraphics[width=\textwidth]{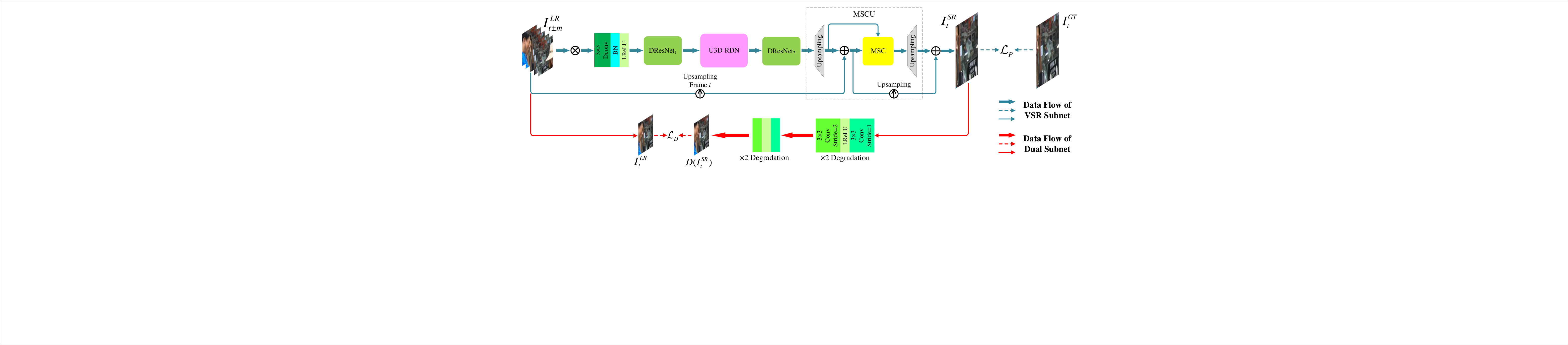}
  \caption{The architecture of the proposed DSMC method. $I_{t\pm m}^{LR}$ denotes the input frames, '$\otimes$' denotes channel-wise concatenation, '$\oplus$' denotes element-wise addition, D($\cdot$) is the dual subnet, and MSC is short for Multi-Stage Communication.}
  \label{fig1}
\end{figure*}

\subsection{VSR Methods Based on Deep Learning}

The earliest application of deep learning to VSR can be traced back to Deep-DE proposed by \citet{liao2015video}. Since then, more advanced VSR methods have been proposed, such as VSRnet \cite{kappeler2016video}, VESPCN \cite{caballero2017real}, SOF-VSR \cite{wang2018learning}, RBPN \cite{haris2019recurrent}, and 3DSRNet \cite{kim20183dsrnet}.

For the VSR methods using 2D convolution, explicit MEMC is widely used and studied. VSRnet used the Druleas algorithm \cite{drulea2011total} to calculate optical flows. In addition, the authors also proposed a filter symmetry mechanism and adaptive motion compensation to accelerate training and reduce the impact of unreliable motion compensated frames. However, the architecture of VSRnet is still relatively simple. Therefore, most recently proposed VSR methods tend to use more complicated explicit MEMC architectures for capturing motion information. \citet{caballero2017real} proposed space converter based on CNN for explicit MEMC. \citet{xue2019video} utilized SpyNet \cite{ranjan2017optical} to estimate optical flows, which were transformed to neighboring frames by a space transformation network \cite{jaderberg2015spatial}.  \citet{kalarot2019multiboot} aligned frames through FlowNet 2.0 \cite{ilg2017flownet}. \citet{wang2018learning} proposed a coarse-to-fine CNN to gradually estimate HR optical flow in three stages.  \citet{sajjadi2018frame} tried to warp the VSR results rather than LR inputs. \citet{haris2019recurrent} presented a back projection module, which is composed of a sequence of alternate encoders and decoders after implicit alignment. And \citet{MAFN} proposed a motion-adaptive feedback cell, which can throw away the redundant information to help useful motion compensation. These methods have achieved better perceptual quality than traditional methods (e.g. bicubic interpolation) and SISR methods with less computational complexity. However, they only work well for videos consisting of scenes with small motion in a short time and less lighting variation. If LR videos are with large motion or significant lighting changes, the performance of explicit MEMC-based methods always degrades greatly.

3D convolution (C3D) can free VSR from explicit MEMC by adding an extra dimensionality. Therefore, C3D-based VSR methods can process videos with various lighting and motion, which enhances their practicability. Existing studies of C3D in VSR are relatively few. \citet{kim20183dsrnet} use extrapolation operation, which can be understood as the padding operation in
temporal domain. \citet{li2019fast} decomposed the general C3D kernels to lower dimensions to reduce computational complexity.  \citet{jo2018deep} proposed a dynamic upsampling filter structure inspired by dynamic filter networks \cite{jia2016dynamic}, which can combine the spatio-temporal information learned by C3D to generate corresponding filters for specific inputs. In order to enhance high-frequency details, an additional network is used to estimate the residual maps. \citet{D3Dnet} proposed a deformable variant of C3D. However, the balance of performance and high computational complexity is still an unsolved challenge which restrains wider application of C3D-based VSR methods.

\section{Proposed Method}

\subsection{Network Architecture}

In this subsection, we give a brief description to the proposed DSMC. It is designed to maximize the communication of various decisive information, so as to maintain excellent robustness in large motion. DSMC includes a VSR subnet (feature extraction and Multi-Stage Communiated Upsampling (MSCU)) and a dual subnet. It uses a center LR frame and $2m$ neighboring frames $I_{t\pm m}^{LR}$ as inputs, and outputs a super-resolved HR frame $I_t^{SR}$. The overall objective function can be formulated as follows:
\begin{equation}
    I_t^{SR}=H_{DSMC}(I)
\label{eq1}
\end{equation}
where $H_{DSMC}(\cdot)$ is our DSMC, and $I$ is a LR frame window defined as
\begin{equation}
    I=[I_{t-m}^{LR}, I_{t-m+1}^{LR},\dots, I_t^{LR}, \dots, I_{t+m-1}^{LR}, I_{t+m}^{LR}]
    \label{eq11}
\end{equation}
where $t$ is the position of the center frame and $m$ is the relative offset of neighboring frames.

The architecture of the proposed DSMC is shown in Figure \ref{fig1}. In details, taking a $\times 4$ VSR task as an example, our model firstly performs deformable convolution on the input consecutive frames for coarse feature extraction. The output feature maps are then processed by a deformable residual network (DResNet) \cite{lei2018temporal} to extract fine spatial information before considering temporal features. Next, the feature maps are input to the U-shaped residual dense network with 3D convolution (U3D-RDN) for dimension reduction and correlation analyzation of spatio-temporal feature. Followed by another DResNet module, the feature maps are sent to a Multi-Stage Communicated Upsampling (MSCU) module. Finally, with the aid of a dual subnet for training, DSMC yields the super-resolved HR frames. It is noted that only the output of the dual subnet, $D(I^{SR}_t)$, and the VSR result, $I^{SR}_t$, are used for the loss computation of DSMC.

In the following subsections, we will give detailed analysis on the motivation and rationality of each module in the proposed DSMC.

\begin{figure}[b]
  \centering
  \includegraphics[width=0.5\textwidth]{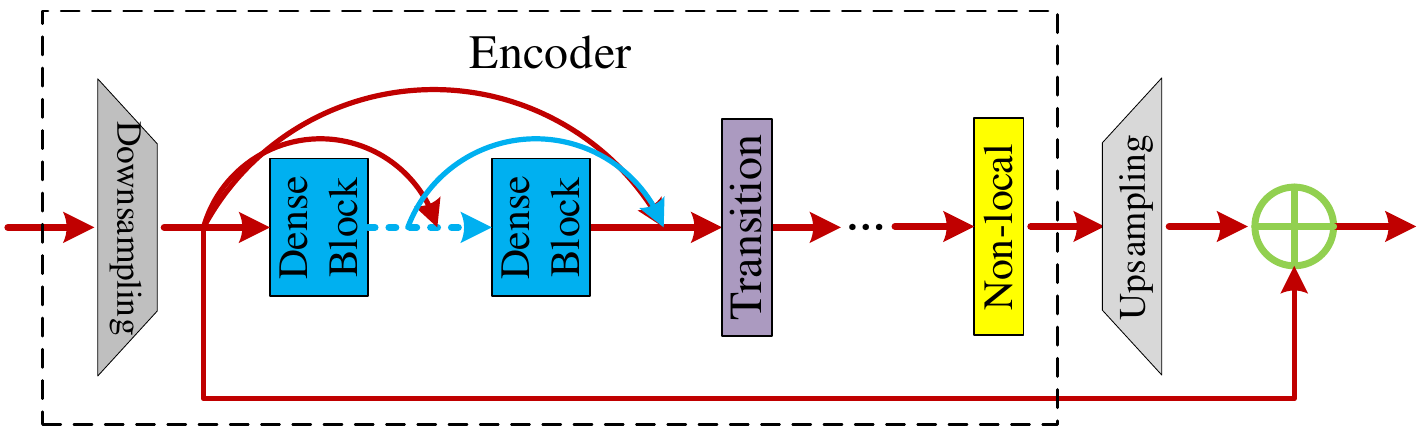}
  \caption{The U3D-RDN module for our DSMC.}
  \label{fig3}
\end{figure}

\subsection{U3D-RDN}

We firstly propose a new U-shaped residual dense network with 3D convolution (U3D-RDN) for our DSMC to achieve both fine implicit MEMC and coarse spatial feature extraction as well as reducing computational complexity, as shown in Figure \ref{fig3}. As it is known, DenseNet \cite{huang2017densely} has achieved sound performance in many deep learning applications. Then \citet{zhang2018residual} proposed a design of residual dense network (ResDenseNet) making full use of the hierarchical features of the LR input in SISR. Unlike the previous work, U3D-RDN is able to execute implicit MEMC without a high computational complexity. It encodes the input features by $\times 2$ downsampling with a $3\times 3$ 2D convolution, and then decodes the residual maps by a sub-pixel convolution upsampling layer after residual learning on the encoding space.

\begin{figure}[ht]
	\centering
	\subfigure[A 3D DenseBlock]{
	    \includegraphics[width=0.23\textwidth]{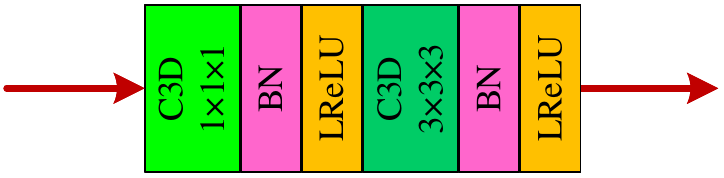}
	}
	\hfil
	\subfigure[A transition layer]{
	    \includegraphics[width=0.16\textwidth]{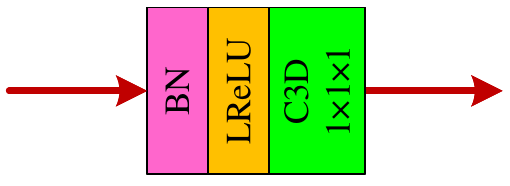}
	}
	\caption{Details of the proposed U3D-RDN module.}
	\label{fig5}
\end{figure}

The proposed module is explained in details below. It consists of $m$ groups of dense blocks with 3D convolution (3D DenseBlock), as shown in Figure \ref{fig5} (a), transition layers in Figure \ref{fig5} (b) among the groups, and a 3D non-local \cite{wang2018non} layer. Our 3D DenseBlock uses 3D convolution (C3D) to avoid the shortage of optical-flow-based methods on videos with large motion, which operates in both spatial and temporal domains of the input features. In a 3D DenseBlock, the $1\times 1\times 1$ C3D is responsible for feature decomposition of the input, while the $3\times 3\times 3$ C3D is for the spatio-temporal feature extraction in the high-dimensional space. These groups of blocks can together establish long-distance dependence and avoid the gradient vanishing problem. The output of the $i$-th group of 3D DenseBlock is expressed as
\begin{equation}
    \mathcal{D}_i(I)=d_i^L([d_i^0(I),d_i^1(I),\dots,d_i^{L-1}(I)])
\label{eq2}
\end{equation}
where $d_i^l(\cdot), l\in (1,\dots,L)$ is the $l$-th 3D DenseBlock in the $i$-th group, and $L$ is the number of the block in each group.

In our U3D-RDN, a transition layer is deployed after each 3D DenseBlock except for the last one. It can halve the temporal dimension of the previous group of cascaded outputs to enhance the learning ability of the module and help realize a deeper network. As it is known, when the number of 3D DenseBlock increases, the dimensions entered by adjacent blocks grow too. Then the learning ability of U3D-RDN can also be enhanced as the network goes deeper. However, the rise of depth also induces a burden on hardware costs. Transition layers can sustain the increase of network depth and relieve the burden of computational costs.

Moreover, the C3D may highlight the shortage of long-distance dependence in CNN due to its synchronous spatio-temporal process. Therefore, before upsampling the output of 3D DenseBlock, a non-local module is adopted to process the feature maps. This non-local module can establish a more relevant dependence on global information.

In summary, the process of U3D-RDN module is given by
\begin{equation}
    \mathcal{R}(I)=\mathcal{N}(\mathcal{D}_m(I)){\uparrow_s}+\mathcal{H}(I)
    \label{eq3}
\end{equation}
where $\mathcal{R}(\cdot)$ is the output, $\mathcal{N}(\cdot)$ is the non-local module, '$\uparrow_s$' is sub-pixel convolution, and $\mathcal{H}$ is the inputs of U3D-RDN.

\subsection{Multi-Stage Communicated Upsampling}

\begin{figure}[t]
	\centering
	\includegraphics[width=0.5\textwidth]{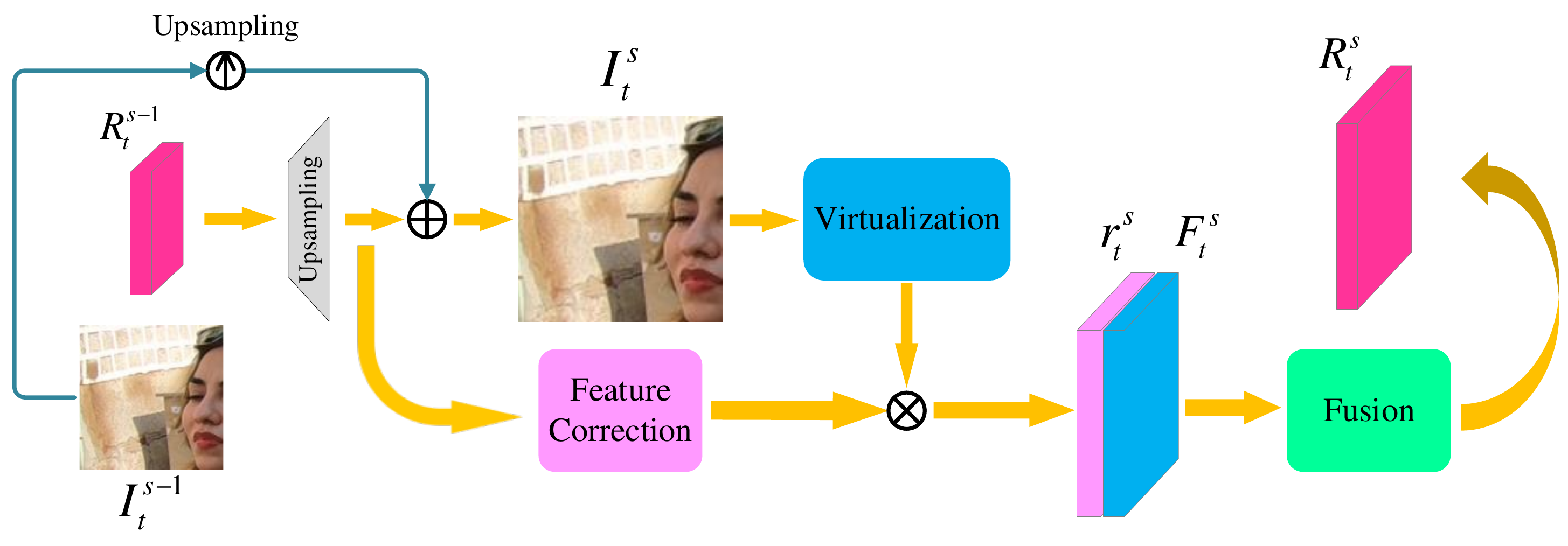}
	\caption{One stage of the proposed MSCU module.}
	\label{fig2}
\end{figure}

We also propose a novel Multi-Stage Communicated Upsampling (MSCU) module to make full use of the prior knowledge in upsampling stages for restoring HR frames. The architecture of MSCU is shown in Figure \ref{fig2}. The curse of dimensionality is considered to be the main factor restricting super-resolution. If the mapping from LR to HR space is directly established, a poor super-resolution algorithm will lack sufficient prior information with the increase of scale factor, causing the distortion of results. In general, there are many sources of prior information. For example, in residual learning, the residual information of the upsampled image after interpolation is used for learning. As shown in \cite{he2016deep}, residual learning usually achieves superior performance. Thus the interpolated image can be a type of prior knowledge, which is not available in a super-resolution network that directly learns HR results.

In our MSCU, in order to take full advantage of prior information, we decompose the upsampling process of VSR to smaller ones. We suggest that each sub scale factor should be in prime to maximize the communication ability of the structure. For example, a $\times 4$ upsampling task can be decomposed into two consecutive $\times 2$ upsampling tasks. Through the sub-outputs, the network will be able to capture the corresponding uncertainties in each stage and try fixing them. The residual maps after upsampling proceed to two branches. One is the feature correction by a ResNet to generate $r_t^s$,  while the other is the channel reduction by a $1\times 1$ convolution and the element-wise addition with the bicubic-upsampled center frame to generate the fixed frame $I_t^s$. Then $I_t^s$ will be virtualized by another $1\times 1$ convolution. The virtual results $F_t^s$ are next channel-wise concatenated to $r_t^s$. The concatenated feature maps are finally fused by a $1\times 1$ convolution to generate residual maps $R_t^s$ for the next stage. The above process can be summarized as follows:
\begin{equation}
    R_t^s=\mathcal{G}([r_t^s,F_t^s])
    \label{eq8}
\end{equation}
where
\begin{equation}
    r_t^s=\mathcal{H}_0((R_t^{s-1})_{\uparrow})
    \label{eq9}
\end{equation}
\begin{equation}
    F_t^s=\mathcal{V}(I_t^s)
    \label{eq10}
\end{equation}
\begin{equation}
    I_t^s=(I_t^{s-1})_{\uparrow}+(R_t^{s-1})_{\uparrow s}
\label{eq12}
\end{equation}
Besides, $s\in\{1, 2, \dots, S-1\}$, $S$ is the number of stages, $\mathcal{G}(\cdot)$ is channel fusion, $\mathcal{H}_0(\cdot)$ is the feature correction network, $\mathcal{V}(\cdot)$ is visulization, and '$\uparrow$' is bicubic upsampling.

Therefore, the equation (\ref{eq1}) is equivalent to
\begin{equation}
    I_t^{SR}=(I_t^{S-1})_{\uparrow}+(R_t^{S-1})_{\uparrow s}
\label{eq5}
\end{equation}

\subsection{Dual Subnet}

We design a novel dual subnet for our DSMC to constrain the solution space. To the best of our knowledge, it is the first design of dual learning in VSR. The architecture of our DSMC with the subnet is shown in Figure \ref{fig1}. The reason that the dual learning mechanism \cite{he2016dual} works is the multi-task connectivity of collaborative operations. It can extend the training constraints to both the outputs and the inputs, which helps to further reduce the solution space and thus makes the network easier to converge.

The purpose of VSR is to map LR frames to their corresponding HR space. Therefore, the dual problem is to restore the degradation results of the VSR outputs as close as possible to the LR target frame. In our proposed dual subnet, we simulate the real image degradation process, which consists of blur, downsampling and noise addition. Mathematically, the SR image $I^{SR}$ and LR image $I^{LR}$ are related by the following degradation model \cite{gu2019blind}:
\begin{equation}
    I^{LR}=({I^{SR}}\otimes k)_{\downarrow s}+n
\label{eq6}
\end{equation}
where $k$ denotes the blur kernel, ${\downarrow s}$ is downsampling process, and $n$ is the added noise. Specifically, the blur and downsampling processes are separately completed by two $3\times 3$ 2D convolutions (C2Ds), while the noise is added to the degraded frame by the bias of the downsampling C2D.

Note that the dual subnet is proposed for helping DSMC converge to a better solution. In our dual learning, the dual loss $\mathcal{L}_D$ is computed between the input frame $I_t^{LR}$ and the output of the dual subnet. Then the total training loss $\mathcal{L}$ of DSMC is composed of two parts: the original loss $\mathcal{L}_P$ from the VSR subnet and $\mathcal{L}_D$, given as follows:
\begin{equation}
    \mathcal{L}=\sum_{t=1}^N  [{\lambda}_1\mathcal{L}_P (I_t^{SR},I_t^{GT})+{\lambda}_2\mathcal{L}_D (D (I_t^{SR}),I_t^{LR})]
\label{eq7}
\end{equation}
where $I_t^{GT}$ is the ground truth, $D$ is the dual subnet with $I_t^{SR}$ as input, ${\lambda}_i,i\in\{1,2\}$ are constants which represent the weight of the two losses, and $N$ is the number of frames.

In fact, we can adopt widely-used Mean Square Error (MSE), Charbonnier (Cb), or perceptual loss for calculating $\mathcal{L}_P$ and $\mathcal{L}_D$. Nevertheless, studies indicate that more strict dual mechanism can better recover the reverse process of the original task. Therefore, we suggest that $\mathcal{L}_P$ and $\mathcal{L}_D$ should use the same loss function to restrain both $I^{LR}$ and $I^{SR}$ to the same distribution. For instance, if $\mathcal{L}_P$ consists of multiple types of losses (e.g. Cb and perceptual loss) with different effects, then $\mathcal{L}_D$ should also include these losses.

\section{Experiments}

\subsection{Datasets and Metrics}

In our experiments, we use REDS \cite{Nah_2019_CVPR_Workshops_REDS} in RGB channels as the training dataset. It contains 240 videos (100 frames for each) for training and 30 videos (100 frames for each) for validation, which are all recorded by GoPro camera in a variety of scenes. As many videos contain large motion, they are challenging for frame alignment. Then we evaluate the performance of our DSMC for VSR tasks on the REDS4 and Vid4 datasets. REDS4 consists of 4 typical videos selected from the REDS validation set, while Vid4 is a benchmark dataset widely used in VSR. All the above datasets are for the $\times 4$ VSR task. To obtain LR videos from the Vid4 benchmark dataset, we downsample HR videos with the scaling factor $\times 4$ by bicubic interpolation. For each dataset except Vid4, we randomly crop frame patches with the LR size of $64\times 64$ as inputs and ground truths (GTs). The mini-batch size during training is set to 8. For data augmentation, we perform randomly horizontal flip and 90 degree rotation. And we evaluate the performance on RGB channels by two most common metrics: peak signal-to-noise ratio (PSNR) and structure similarity index (SSIM).

\begin{table*}[htbp]
    \centering
    \caption{PSNR/SSIM results of different methods on the REDS4 dataset for scale factor $\times 4$. $Params.$ is short for the number of parameters ($\times 10^6$). All the metrics are compared on RGB channels. Note that the the best results are in {\color{red}{\textbf{red boldface}}} and the second best results are in {\color{blue}{\textbf{blue boldface}}}.}
    \label{tab2}
    \renewcommand\arraystretch{1.35}
    \resizebox{\textwidth}{!}{
    \begin{tabular}{|l|c|c|c|c|c|c|c|c|c|c|}
    \hline
        Clip Name & Bicubic & SRCNN & DBPN & VESPCN & SOF-VSR & 3DSRnet & RBPN & DRVSR & FRVSR & DSMC (ours) \\ \hline
        Clip\_000 & 24.83/0.8076  & 25.88/0.8443  & \color{blue}{\textbf{27.24/0.8831}} & 26.57/0.8649  & 26.82/0.8722  & 25.30/0.8230  & 27.10/0.8785  & 26.53/0.8640  & 26.98/0.8776  & \color{red}{\textbf{27.56/0.8934}}  \\
        Clip\_011 & 21.60/0.7061  & 22.61/0.7628  & 23.12/0.7859  & 22.89/0.7753  & 22.97/0.7785  & 22.16/0.7399  & 23.04/0.7812  & 22.84/0.7722  & \color{blue}{\textbf{23.23/0.7901}}  & \color{red}{\textbf{23.57/0.8070}}  \\
        Clip\_015 & 21.12/0.6687  & 21.88/0.7236  & 22.46/0.7543  & 22.08/0.7378  & 22.22/0.7445  & 21.66/0.7116  & 22.28/0.7479  & 22.10/0.7379  & \color{blue}{\textbf{22.63/0.7676}}  & \color{red}{\textbf{23.11/0.7897}}  \\
        Clip\_020 & 27.33/0.8412  & 27.91/0.8584  & \color{blue}{\textbf{28.29/0.8673}}  & 28.18/0.8648  & 28.21/0.8668  & 27.88/0.8554  & 28.25/0.8671  & 28.15/0.8643  & 28.25/0.8671  & \color{red}{\textbf{28.68/0.8810}}  \\
        Average & 23.72/0.7559  & 24.57/0.7973  & {\color{blue}{\textbf{25.28}}}/0.8227 & 24.93/0.8107  & 25.05/0.8155  & 24.25/0.7825  & 25.17/0.8187  & 24.90/0.8096  & 25.27/\color{blue}{\textbf{0.8256}}  & \color{red}{\textbf{25.73/0.8428}}  \\ \hline
        Params. & - & 0.07M & 10.42M & 0.88M & 1.71M & 0.11M & 14.51M & 2.17M & 2.81M & 11.58M \\ \hline
    \end{tabular}}
\end{table*}
\begin{figure*}[ht]
    \centering
    \subfigure{
        \begin{minipage}[b]{\textwidth}
        \includegraphics[width=\textwidth]{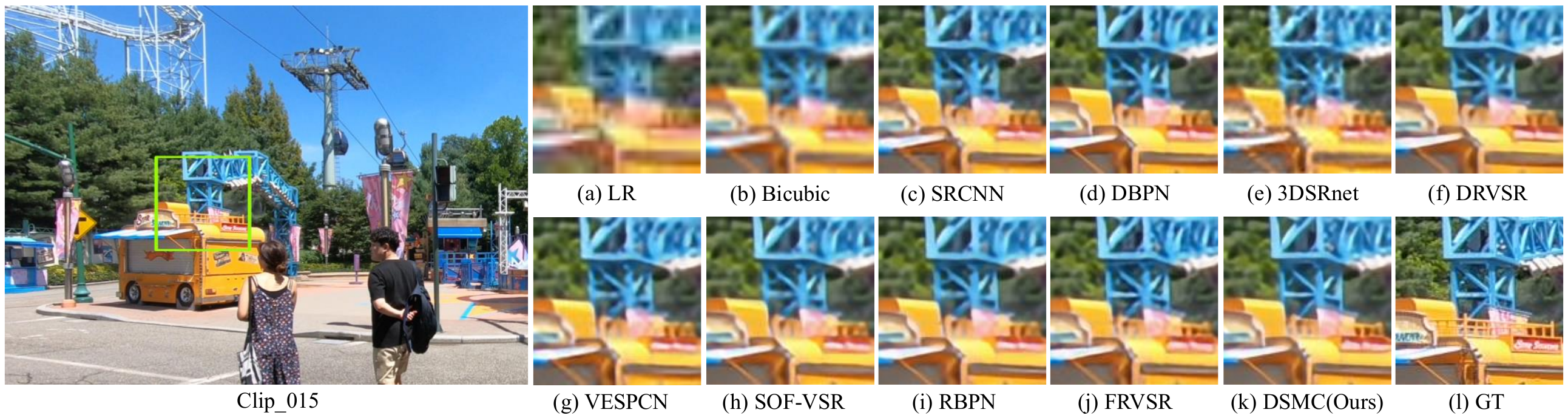}
        \end{minipage}
    }
    \caption{Qualitative results by different methods on the REDS4 dataset.}
    \label{fig6}
\end{figure*}

\subsection{Implementation Details}

Our DSMC uses 5 consecutive frames as the inputs. For the beginning and the end frames that do not have enough frames in the window, we pad with different Gaussian-blured center frames with $radius=0.1d$ ($d$ is the distance from center to the padding) to simulate the focusing and defocusing of camera. $\alpha$ is set to 0.1 in the LReLUs of the whole network. In U3D-RDN, the number of the groups of 3D DenseBlocks $m$ is set to 4, and the sizes which represent the number of 3D DenseBlocks in each group are set to (2, 6, 6, 3), respectively. Our model is trained by the Adam optimizer with the momentum parameter $\beta_1=0.9$. The loss function is defined in Equation (\ref{eq7}), where ${\lambda}_1=1$ and ${\lambda}_2=0.1$, $\mathcal{L}_P$ and $\mathcal{L}_D$ are of the Cb loss \cite{cui2019class} and perceptual loss \cite{johnson2016perceptual}, where the weight of perceptual loss $\lambda_p$ is set to 0.1. The initial learning rate is set to $2e^{-4}$. We apply the PyTorch framework to implement the proposed network and train it on a desktop computer with 2.40GHz Intel Xeon CPU E5-2640 v4, 64GB RAM, and a NVIDIA GTX 1080Ti GPU.

\subsection{Model Analysis}

\subsubsection{Comparison with state-of-the-art methods}

We implement 8 state-of-the-art algorithms for comparison including SRCNN \cite{dong2015image}, DBPN \cite{haris2018deep}, DRVSR \cite{tao2017detail}, VESPCN \cite{caballero2017real}, SOF-VSR \cite{wang2018learning}, RBPN \cite{haris2019recurrent}, FRVSR \cite{sajjadi2018frame} and 3DSRnet \cite{kim20183dsrnet}. The first 2 methods are SISR, while others are VSR methods. For fair comparison, for each algorithm, we control number of iterations to around 30,000 during the training process. Then we choose 4 representative videos (400 frames in total) from the validation set of REDS for validation, which is together named REDS4. We crop the original REDS4 to $32\times 32$ frame patches without augmentation to create 4 smaller consecutive video patches.

Table \ref{tab2} shows quantitive comparisons for $\times 4$ VSR on REDS4. Obviously, our DSMC gains an average PSNR/SSIM of 25.73/0.8428 and outperforms other methods in all the settings. It is clear that the PSNR and SSIM of optical-flow-based methods (e.g. VESPCN and RBPN) are relatively low, which verifies that they are subject to large motion in videos and produce inferior performance. Qualitative comparisons are shown in Figure \ref{fig6}. It can be seen that the outputs of DSMC show fewer artifacts and better perceptual quality, which confirms the effectiveness of our DSMC.

\begin{figure*}[ht]
    \centering
    \subfigure{
        \begin{minipage}[b]{\textwidth}
        \includegraphics[width=\textwidth]{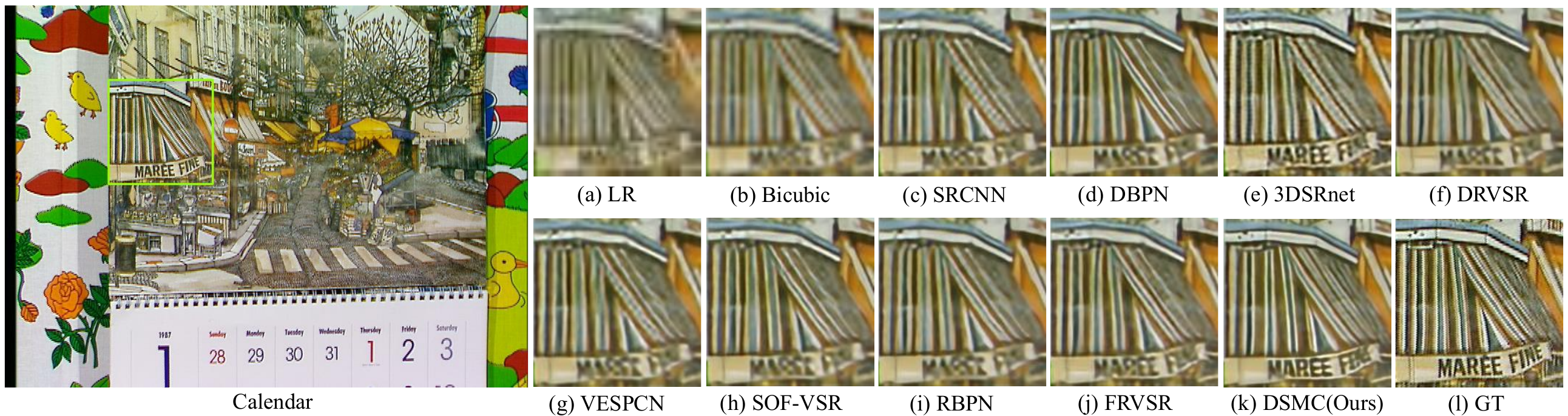}
        \end{minipage}
    }
    \caption{Qualitative results by different methods on the Vid4 dataset.}
    \label{fig7}
\end{figure*}

Moreover, we also test our model on the Vid4 benchmark dataset. The quantitive results are listed in Table \ref{tab3}. Our DSMC still yields the highest PSNR and SSIM, while 3DSRnet yields the second highest. Besides, we find that performance of 3DSRnet has reduced when the dataset changes from Vid4 to REDS4. We argue that this is likely due to the relatively simple structure, which reduces its ability of modeling and handling with videos with large motion. On the other hand, the generalization ability of our DSMC is verified. Qualitative comparisons are shown in Figure \ref{fig7}.

\begin{table*}[ht]
    \centering
    \caption{PSNR/SSIM results of different methods on the Vid4 dataset for scale factor $\times 4$. All the metrics are compared on RGB channels. Note that the best results are in {\color{red}{\textbf{red boldface}}} and the second best results are in {\color{blue}{\textbf{blue boldface}}}.}
    \label{tab3}
    \renewcommand\arraystretch{1.35}
    \resizebox{\textwidth}{!}{
    \begin{tabular}{|l|c|c|c|c|c|c|c|c|c|c|}
    \hline
        Clip Name & Bicubic & SRCNN & DBPN & VESPCN & SOF-VSR & 3DSRnet & RBPN & DRVSR & FRVSR & DSMC (ours) \\ \hline
        Calendar & 18.98/0.6629 & 19.64/0.7184 & 20.23/0.7470 & 20.05/0.7401 & 20.06/0.7456 & \color{blue}{\textbf{20.76/0.7818}} & 20.09/0.7468  & 20.21/0.7526 & 20.18/0.7507 & \color{red}{\textbf{21.10/0.7981}}  \\
        City & 23.76/0.7275 & 24.17/0.7616 & 24.54/0.7843 & 24.48/0.7796  & 24.45/0.7795  & \color{blue}{\textbf{25.32/0.8264}}  & 24.49/0.7808  & 24.60/0.7882  & 24.74/0.7970  & \color{red}{\textbf{25.54/0.8413}}  \\
        Foliage & 22.20/0.6920 & 22.88/0.7409  & 23.20/0.7580 & 23.13/0.7526  & 23.18/0.7553  & \color{blue}{\textbf{23.96/0.7924}}  & 23.12/0.7510  & 23.23/0.7582 & 23.42/0.7684  & \color{red}{\textbf{23.94/0.7954}}  \\
        Walk & 24.94/0.8650  & 26.34/0.8982 & 27.09/\color{blue}{\textbf{0.9135}} & 26.88/0.9079  & 26.90/0.9092  & {\color{blue}{\textbf{27.45}}}/0.9042  & 26.88/0.9095  & 26.89/0.9085  & 27.04/0.9111  & \color{red}{\textbf{27.92/0.9263}}  \\
        Average & 22.47/0.7369 & 23.26/0.7798  & 23.77/0.8007 & 23.63/0.7951  & 23.65/0.7974  & \color{blue}{\textbf{24.37/0.8262}}  & 23.65/0.7970  & 23.73/0.8019  & 23.85/0.8068  & \color{red}{\textbf{24.63/0.8403}}  \\ \hline
        Params. & - & 0.07M & 10.42M & 0.88M & 1.71M & 0.11M & 14.51M & 2.17M & 2.81M & 11.58M \\ \hline
    \end{tabular}}
\end{table*}

\begin{table}[t]
    \centering
    \caption{Ablation studies on our DSMC. $Comm.$ is short for communication, and $MSCU_{(\cdot)}$ denotes the submodule of MSCU. The most effective module and the second one are in {\color{red}{\textbf{red boldface}}} and {\color{blue}{\textbf{blue boldface}}}, respectively. }
    \label{tab4}
    \renewcommand\arraystretch{1.35}
    \resizebox{0.48\textwidth}{!}{
    \begin{tabular}{lccc}
    \hline
        Ablation & PSNR/SSIM & $\Delta$\\ \hline
        Bicubic & 23.72/0.7559 & -2.01/-0.0869\\ \hline
        w/o MSCU & 20.48/0.6264 & {\color{red}{\textbf{-5.25/-0.2164}}}\\
        w/o MSCU$_{ResNet}$ & 24.81/0.8210 & -0.92/-0.0218\\
        w/o MSCU$_{Comm.}$ & 22.30/0.6853 & -3.43/-0.1575\\
        w/o MSCU$_{ResNet+Comm.}$ & 21.77/0.6711 & {\color{blue}{\textbf{-3.96/-0.1717}}}\\
        w/o U3D-RDN & 22.64/0.7093 & -3.09/-0.1335\\
        w/o U3D-RDN$_{Non\_local}$ & 25.59/0.8378 & -0.14/-0.0050\\ \hline
        \textbf{DSMC (Baseline)} & {\textbf{25.73/0.8428}} & {\textbf{0.00/0.0000}}\\ \hline
    \end{tabular}}
\end{table}

\begin{table}[t]
	\centering
	\caption{FLOPs and parameter numbers of traditional 3D-RDN (without U-shaped structure) and the proposed U3D-RDN. Note that the mini-batch size of the test data is 1.}
	\label{tab5}
	\renewcommand\arraystretch{1.25}
	\begin{tabular}{lccc}
		\hline
		Module & FLOPs & Params. \\ \hline
		3D-RDN & 509.17G & 2.48M \\
		U3D-RDN & 129.18G & 2.52M \\ \hline
	\end{tabular}
\end{table}

\begin{table}[t]
    \centering
    \caption{Ablation studies on the dual loss. $Perc.$ is short for perceptual loss. Noted that the best results are in {\color{red}{\textbf{red boldface}}} and the second best results are in {\color{blue}{\textbf{blue boldface}}}.}
    \label{tab6}
    \renewcommand\arraystretch{1.1}
    \resizebox{0.45\textwidth}{!}{
    \begin{tabular}{ccccc}
    \hline
        Cb & Perc. & Dual Cb & Dual Perc. & PSNR/SSIM \\ \hline
        $\checkmark$ &  &  &  & 25.70/0.8408  \\
        $\checkmark$ &  & $\checkmark$ &  & {\color{blue}{\textbf{25.71/0.8420}}} \\
        $\checkmark$ & $\checkmark$ &  &  & 25.67/0.8413 \\
        $\checkmark$ & $\checkmark$ & $\checkmark$ & $\checkmark$ & {\color{red}{\textbf{25.73/0.8428}}} \\ \hline
    \end{tabular}}
\end{table}

\subsubsection{Ablation studies}

In order to verify the effectiveness of each module, we do several ablation studies on important modules of DSMC. We use video clips in the validation set of REDS for test. The experimental results are shown in Table \ref{tab4}. In details, we ablate the MSCU module by replacing it with a $\times 4$ upsampling layer, and the communication mechanism by simply executing two $\times 2$ upsampling.

It is seen that MSCU poses the most significant influence on the performance of DSMC, while the join of feature correction network (ResNet) and communication mechanism is the key factor that determines the ability of MSCU. Particularly, U3D-RDN also plays an important role, which indicates that it has strong ability for modeling large motion in videos. Furthermore, we do an extra study to demonstrate the effectiveness of U3D-RDN on reducing computational complexity, as shown in Table \ref{tab5}. Obviously, the proposed U3D-RDN has reduced 74.6\% of the computational cost with only 0.04M additional parameters.

In addition, we also analyze the effectiveness of the dual subnet in Table \ref{tab6}. We train 4 models of DSMC with different loss functions. Among them, the Cb and perceptual loss functions are used for the VSR subnet, and the dual Cb and dual perceptual losses are for the dual subnet. The results show that when both the VSR subnet and the dual subnet use Cb and perceptual losses, the values of PSNR and SSIM are the best. It confirms the strict dual learning mechanism. Moreover, the second best performance is obtained when both the VSR and the dual subnet employ only Cb loss. Additionally, when there is no dual subnet (e.g. only Cb loss exists), the performance of DSMC degrades, which indicates the role of dual subnet. It is noted that the contribution brought by the dual subnet is less obvious than the proposed VSR subnet according to Tables \ref{tab4} and \ref{tab6}, which demonstrate that a reasonable design of a VSR network is of more vital importance than an ingenious training strategy.

\section{Conclusion}

In this paper, we proposed a novel Video Super-Resolution Network with Dual Subnet and Multi-Stage Communicated Upsampling (DSMC). We designed a new U-shaped residual dense network with 3D convolution (U3D-RDN) for our DSMC, which can achieve both fine implicit motion estimation and motion compensation (MEMC) and coarse spatial feature extraction as well as reducing the computational complexity. Moreover, we proposed a Multi-Stage Communicated Upsampling (MSCU) module for helping utilize intermediate information during upsampling, and a dual subnet which can enhance the generalization ability. Extensive experimental results confirmed the effectiveness of our method in processing videos with large motion for $\times 4$ VSR tasks. Additionally, we did ablation studies on important modules of our DSMC. They indicated that U3D-RDN and MSCU are the key modules that affect the performance of DSMC. Meanwhile, the proposed dual loss can help DSMC converge to a better solution. We believe that our work can provide a better viewing experience for extreme sports fans and outdoor photographers. In the future, we will improve our design for large-scale video super-resolution.

\section*{Acknowledgments}
This work was supported by the National Natural Science Foundation of China (Nos.\ 61976164, 61876220, 61876221, 61836009, U1701267, and 61871310), the Project supported the Foundation for Innovative Research Groups of the National Natural Science Foundation of China (No.\ 61621005), the Major Research Plan of the National Natural Science Foundation of China (Nos.\ 91438201 and 91438103), the Program for Cheung Kong Scholars and Innovative Research Team in University (No.\ IRT\_15R53), the Fund for Foreign Scholars in University Research and Teaching Programs (the 111 Project) (No.\ B07048), the Science Foundation of Xidian University (Nos.\ 10251180018 and 10251180019), the National Science Basic Research Plan in Shaanxi Province of China (Nos.\ 2019JQ-657 and 2020JM-194), and the Key Special Project of China High Resolution Earth Observation System-Young Scholar Innovation Fund.

\bibliography{An.bib}

\end{document}